\documentclass{article}

\usepackage{arxiv}

\usepackage{amsmath}
\usepackage[backend=bibtex, style=numeric]{biblatex}  
\usepackage{nicefrac}
\usepackage{listings} 
\usepackage[usenames,dvipsnames]{xcolor}
\usepackage{graphicx}
\usepackage{multicol,lipsum}
\usepackage{tabularx}

\usepackage[utf8]{inputenc} 
\usepackage[T1]{fontenc}   
\usepackage{hyperref}      
\usepackage{url}           
\usepackage{booktabs}      
\usepackage{amsfonts}           
\usepackage{microtype}     	

\usepackage{draftwatermark}
\SetWatermarkText{PREPRINT}
\SetWatermarkScale{1}
\definecolor{wm}{gray}{0.95}
\SetWatermarkColor{wm}

\newcommand{\papertitle}{NeuralFMU: Presenting a workflow for integrating hybrid NeuralODEs into real world applications}

\newcommand{\reffig}[1]{Figure~\ref{fig:#1}}
\newcommand{\reffigs}[3]{Figures~\ref{fig:#1},~\ref{fig:#2}~and~\ref{fig:#3}}
\newcommand{\reftab}[1]{Table~\ref{tab:#1}}
\newcommand{\refsec}[1]{Section~\ref{sec:#1}}
\newcommand{\refsecs}[2]{Sections~\ref{sec:#1}~and~\ref{sec:#2}}

\newcommand{\traincycle}[0]{CADC (Road)}
\newcommand{\testcycle}[0]{WLTC (class 2)}
\newcommand{\batchlen}[0]{$100\,s$}
\newcommand{\numbatches}[0]{11}
\newcommand{\numepochs}[0]{18}
\newcommand{\trainlength}[0]{$\approx 5\,hours$}

\newcommand{\libfmi}{{\emph{FMI.jl}}}
\newcommand{\libfmiflux}{{\emph{FMIFlux.jl}}}
\newcommand{\libflux}{{\emph{Flux.jl}}}

\newcommand{\libdiffeqflux}{{\emph{DiffEqFlux.jl}}}
\newcommand{\libmodia}{{\emph{Modia.jl}}}

\newcommand{\urlfmi}{{\url{https://github.com/ThummeTo/FMI.jl}}}
\newcommand{\urlfmiflux}{{\url{https://github.com/ThummeTo/FMIFlux.jl}}}

\newcommand{\urlflux}{\url{https://fluxml.ai/Flux.jl/stable/}}

\newcommand{\urldiffeqflux}{\url{https://github.com/SciML/DiffEqFlux.jl}}
\newcommand{\urlmodia}{\url{https://github.com/ModiaSim/Modia.jl}}

\newcommand{\mycode}[1]{{\small\texttt{#1}}}

\let\vec\mathbf

\usepackage{multirow}

\usepackage{acronym}

\renewcommand{\headeright}{Preprint}
\renewcommand{\undertitle}{Preprint}
\renewcommand{\shorttitle}{~}

\hypersetup{%
  pdftitle  = {\papertitle},
  pdfauthor = {Tobias Thummerer, Johannes Stoljar, Lars Mikelsons},
  pdfkeywords = {NeuralFMU; FMU; Functional Mock-up Unit; NeuralODE; hybrid model; FMI; vehicle longitudinal dynamics model; PhysicsAI; scientific machine learning}
}

\begin{document}

\title{\papertitle}
\author{%
	Tobias Thummerer \\
	Chair of Mechatronics\\
	Augsburg University \\
	\texttt{tobias.thummerer@uni-a.de} 
	\And 
	Johannes Stoljar \\
	Chair of Mechatronics\\
	Augsburg University \\
	\texttt{johannes.stoljar@uni-a.de} 
	\And 
	Lars Mikelsons \\
	Chair of Mechatronics\\
	Augsburg University \\
	\texttt{lars.mikelsons@uni-a.de}
}
\date{}
\maketitle

\begin{abstract}
The term \emph{NeuralODE} describes the structural combination of an \ac{ANN} and a numerical solver for \acp{ODE}, the former acts as the right-hand side of the \ac{ODE} to be solved. This concept was further extended by a black-box model in the form of a \ac{FMU} to obtain a subclass of NeuralODEs, named \emph{NeuralFMUs}. The resulting structure features the advantages of first-principle and data-driven modeling approaches in one single simulation model: A higher prediction accuracy compared to conventional \acp{FPM}, while also a lower training effort compared to purely data-driven models. We present an intuitive workflow to setup and use NeuralFMUs, enabling the encapsulation and reuse of existing conventional models exported from common modeling tools. Moreover, we exemplify this concept by deploying a NeuralFMU for a consumption simulation based on a \ac{VLDM}, which is a typical use case in automotive industry. Related challenges that are often neglected in scientific use cases, like real measurements (e.g. noise), an unknown system state or high-frequent discontinuities, are handled in this contribution. For the aim to build a hybrid model with a higher prediction quality than the original \ac{FPM}, we briefly highlight two open-source libraries: \libfmi{} for integrating \acp{FMU} into the Julia programming environment, as well as an extension to this library called \libfmiflux{}, that allows for the integration of \acp{FMU} into a neural network topology to finally obtain a \emph{NeuralFMU}.\acresetall
\end{abstract}


\section{Introduction}
\emph{Hybrid modeling} describes on the one hand the field of research in machine learning that focuses on the fusion of \acp{FPM}, often in the form of symbolic differential equations, and machine learning structures like \acp{ANN}. On the other hand, in the field of \acp{ODE}, \emph{hybrid models} name the piece-wise concatenation of continuous models over time to obtain a discontinuous model, of which the numerically simulated solutions may not be continuously differentiable over time. In this article, we present a workflow concerning both interpretations of \emph{hybrid modeling} by integrating custom, discontinuous simulation models and \acp{ANN} into a discontinuous NeuralFMU. For illustration, we use the example of learning a friction model for an industry typical automotive consumption simulation based on a \ac{VLDM}.

In the following, the term \emph{hybrid model} is used to identify a model based on the combination of a \ac{FPM} and \ac{ML}, whereas the concatenation of multiple continuous systems is referred to as \emph{discontinuous model}. 

\clearpage
\subsection{State-of-the-art: Hybrid modeling}
As part of research applications, the structural integration of physical models inside \ac{ML}-topologies like \acp{ANN} is a topic growing attention. One approach for hybrid modeling is the integration of the \ac{FPM} into the \ac{ML} process by evaluating the forward propagation of the physical model as part of the loss function during training, namely \acp{PINN} \cite{Raissi:2019}. In contrast, our method focuses on the structural integration of \acp{FPM} into the \ac{ANN}/\ac{ODE} itself and \emph{not} only in the cost function, allowing much more flexibility with respect to what can be learned and influenced. However, it is also possible to build and train \acp{PINN} with the presented libraries. Another approach uses \acp{BNSDE}, which applies model selection together with \ac{PAC} Bayesian bounds during the \ac{ANN} training to improve hybrid model accuracy on basis of noisy prior knowledge \cite{Haussmann:2021}. Besides, \acp{DARN} can also be used to model physical systems. Similar to \acp{RNN}, the output of the last network inference is fed back into the neural network itself as input \cite{Karol:2013}. Different from \acp{RNN}, this feedback is not modeled as neural network state, but a simple feed-forward connection, and thus a \ac{DARN} can be trained as conventional feed-forward network with all related simplifications and benefits. Finally, the combination of symbolic \acp{ODE} and object-orientated modeling languages like \emph{Modia} is a promising and emerging research field too, because of the benefits of acausal modeling \cite{Bruder:2021}. For a general overview on the growing field of hybrid modeling see e.g. \cite{Willard:2020} or \cite{Rai:2020}.\\
With regards to learning system dynamics (the right-hand side of a differential equation), the structural integration of algorithmic numerical \ac{ODE}-solvers into \acp{ANN} leads to significant improvements in performance, memory cost and numerical precision in comparison to the use of residual neural networks \cite{Chen:2018}, while offering a new range of possibilities, e.g. fitting data observed at irregular time steps \cite{Innes:2019}. This integration of a numerical solver for \acp{ODE} into an \ac{ANN} is known as \emph{NeuralODE}, which is further introduced in \refsec{neuralode}. Probably the most mentioned point of criticism regarding NeuralODEs is the difficult transfer to real world applications for the following reasons:
\begin{itemize}
	\item	Real world models from common modeling tools are in general not available as symbolic \acp{ODE};
	\item   Neural\ac{ODE} training tends to converge in local minima;
	\item	Neural\acp{ODE} training often takes a considerable amount of calculation time.
\end{itemize}
Whereas the tendency to early converge to local minima and long training times can be tackled by different techniques (s. \refsecs{initialization}{training}), the major technical challenge that hinders hybrid modeling in industrial applications remains: \acp{FPM} are modeled and simulated inside closed tools. \ac{ML} features, the foundation to allow for hybrid modeling, are missing in such tools and a seamless interoperability to \ac{ML}-frameworks is not given. For example, to train the data-driven parts of hybrid models, determination of the loss function gradient through the \acp{ANN} and the \acp{FPM} themselves is needed. This requires different high-performance sensitivity algorithms like \ac{AD}. \ac{ML}-frameworks on the other hand, provide these abilities. To build high performing hybrid models, an interface between these two application worlds is needed.

\subsection{Preliminary work: NeuralFMUs}
In a preliminary publication \cite{ThummererModelica:2021}, we started facing that issue and extended the concept of Neural\acp{ODE} by adding \acp{FPM} in the form of \acp{FMU} into this topology (s. \refsec{neuralode}). The resulting subclass of hybrid Neural\acp{ODE}, called \emph{NeuralFMUs}, can be seen as the injection of system knowledge in the form of a \ac{FPM} into the \ac{ANN} model, equivalently the right-hand side of the \ac{ODE}. This not only enhances the prediction quality, but also significantly reduces the amount of necessary training data, because \emph{only} missing physical effects need to be learned. For example, the original \acp{FPM} were outperformed in terms of computational performance \cite{ThummererICMSquare:2021} and prediction accuracy \cite{ThummererModelica:2021}, with trained only on data gathered by a single, short part of simulation trajectory. Further, the integration of a \ac{FPM} can strongly enhance the extrapolation capabilities of the hybrid model compared to conventional pure data-driven models.
In this article, we want to follow up these publications \cite{ThummererModelica:2021, ThummererICMSquare:2021} by providing a workflow and results not only for a synthetic example, but also a real world application in the form of an energy consumption simulation based on a \ac{VLDM}. The following challenges, which are common in industrial applications but are often neglected for scientific experiments, are overcome as part of this publication: Discontinuity, closed-loop controllers inside of the system, an unknown system state and typical measurement errors (e.g. noise). 

This article is further structured into three sections: A brief introduction to the used standards, methods, the corresponding software libraries \libfmi{} and \libfmiflux{} and the \ac{VLDM}. This is followed by the presentation of results of the example use case handling a Neural\ac{FMU} setup and training and finally closed by a conclusion with future outlook.

\section{Materials and Methods}
In this section, a short overview over the used standards, software and methods is given. On foundation of these, a workflow for the setup of NeuralFMUs in real applications is given and methods of initialization and training are elaborated. Finally, the \ac{VLDM}, the \ac{FPM} for the considered example use case, is introduced.
\subsection{\ac{FMI}}
\begin{figure}[h!]
	\centering
	\includegraphics[height=7.0cm]{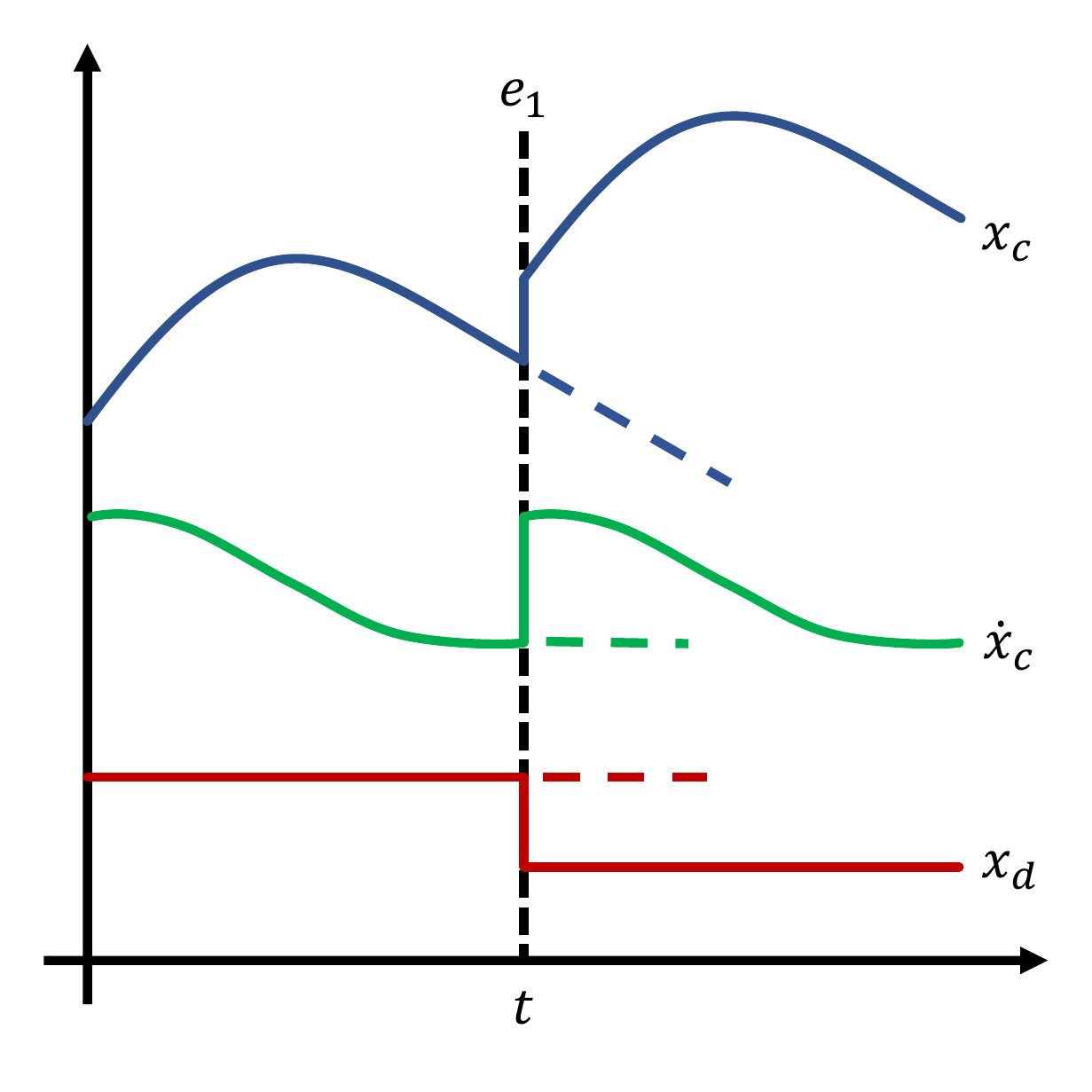}
	\caption{Exemplified simulation of a \ac{ME}-\ac{FMU} with events (hybrid \ac{ODE}). The single (piecewise) continuous system state $x_c$ (blue) and state derivative $\dot{x}_c$ may depend on a discrete system state $x_d$ (red), which is unknown in general for \acp{FMU}. Not handling events like $e_1$ (black-dashed), leads to wrong system values (blue-, green-, red-dashed), because the system state is not updated properly.}
	\label{fig:events}
\end{figure}

The \ac{FMI} standard\footnote{\url{https://fmi-standard.org/}} allows for the distribution and exchange of models in a standardized format independent of the modeling tool. The interface standard counts three version releases, the most popular version is 2.0 \cite{FMI:2020}, the successor version 3.0 \cite{FMI:2022} was released in May 2022. An exported model container that fulfills the \ac{FMI} requirements is called \ac{FMU}. \acp{FMU} can be used in other simulation environments or even as part of entire \aclp{CS} or \ac{SSP} \cite{SSP:2019}. \acp{FMU} are subdivided into three major classes: \ac{ME}, \ac{CS} and new in FMI~3.0 \ac{SE}. The possible use cases depend on the \ac{FMU}-type and the availability of standardized and optionally implemented \ac{FMI} functions. Most relevant for the considered use case are \ac{ME}-\acp{FMU}, because this type allows for manipulation and extension of system dynamics before the numerical integration.\\
To optimize simulation performance, fast physical effects like the change from stick- to slide-friction or the firing of an electrical diode are often modeled in a discontinuous way. This means that the expressions of the right-hand side of the \ac{ODE} model may change depending on the current system state and time, whereas this transition is discrete. Inside \ac{FMI}, this means \ac{ME}-\acp{FMU} may contain state- or time-dependent discontinuities, which are triggering so called \emph{events}. The actual event time point, the instant at which the equations and/or the state of the model is modified, is defined by a predefined time point itself (time-events) or the zero-crossing of a scalar value (state-events), also called the \emph{event indicator}. For a detailed view on the event definition and handling, see \cite{FMI:2022}. Basically, any \ac{ME}-\ac{FMU} with continuous state $\vec{x}_c$, discrete state $\vec{x}_d$ and time- and/or state-events can be seen as a hybrid \ac{ODE}, as in \reffig{events}. Continuous states may change in time, while discrete states can only change their value at event instances.

\clearpage
\subsection{NeuralODEs \& NeuralFMUs}\label{sec:neuralode}
\begin{figure}[h!]
	\centering
	\includegraphics[height=3cm]{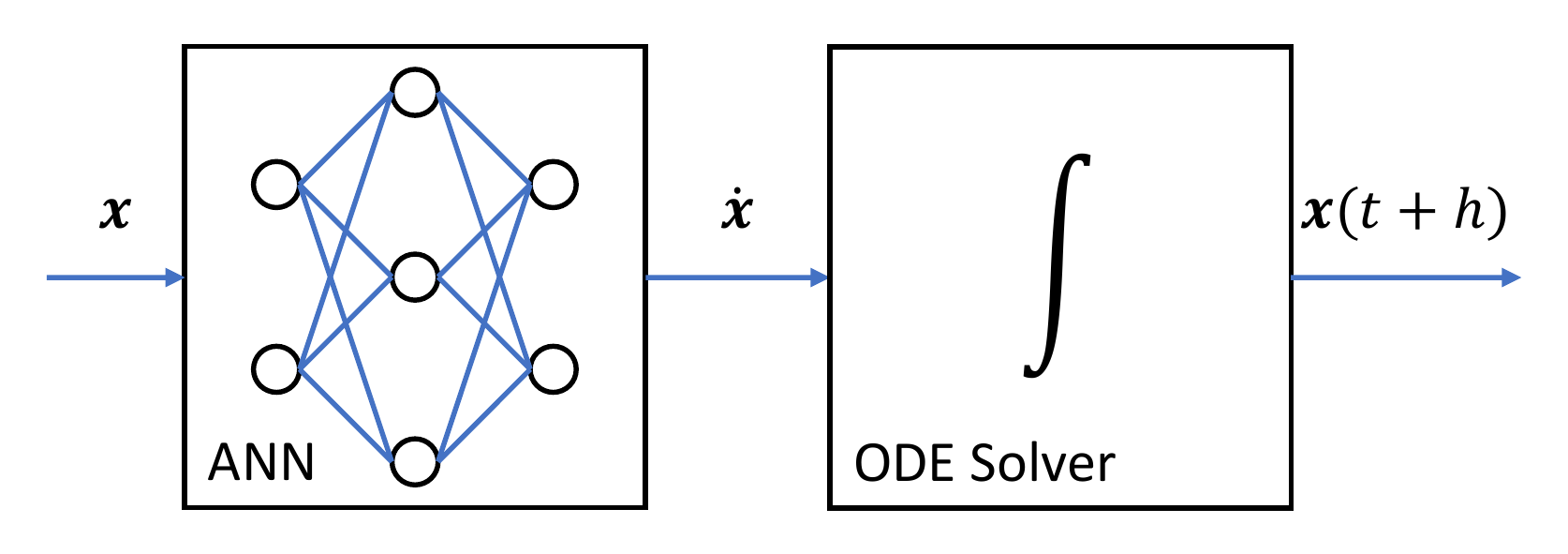} 
	\caption{Topology of a NeuralODE consisting of a feed-forward \ac{ANN} and a numerical \ac{ODE}-solver: The current system state $\vec{x}$ is passed to the \ac{ANN}, based on that the state derivative $\dot{\vec{x}}$ is calculated and integrated into the next system state $\vec{x}(t+h)$ by the \ac{ODE}-solver with time step size $h$.\label{fig:neuralode}}
\end{figure}

Neural\acp{ODE} are defined by the structural combination of an \ac{ANN} and a numerical \ac{ODE}-solver, see figure \ref{fig:neuralode}. As a result, the \ac{ANN} acts as the right-hand side of an \ac{ODE}, whereas solving of this \ac{ODE} is performed by a conventional \ac{ODE}-solver. If external requirements (tolerance or stiffness) change, the \ac{ODE}-solver can be easily replaced by another one. The scientific contribution at this point is not only the idea of this subdivision, but also, more importantly, a concept to allow training this topology on a target solution for the \ac{ODE}. This requires propagation of the parameter sensitivities of the \ac{ANN} through the \ac{ODE}-solver \cite{Chen:2018}. \libdiffeqflux{}\footnote{\urldiffeqflux{}} is already available as a ready-to-use library for building and training NeuralODEs in the Julia programming language \cite{Rackauckas:2019}.

\begin{figure}[h!]
	\centering
	\includegraphics[height=3cm]{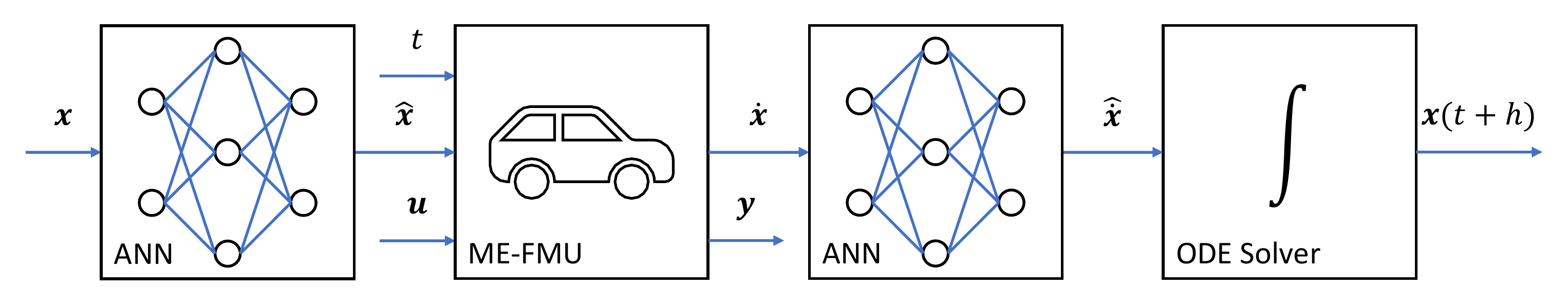} 
	\caption{Topology of a \ac{ME}-NeuralFMU (example) consisting of two feed-forward \acp{ANN}, a \ac{ME}-\ac{FMU} and a numerical \ac{ODE} solver: The current system state $\vec{x}$ is passed to a (state) \ac{ANN}, based on that a manipulated system state $\hat{\vec{x}}$ is calculated and passed to the \ac{ME}-\ac{FMU}. There, the system state derivative $\dot{\vec{x}}$ is computed and manipulated by another (derivative) \ac{ANN} into the changed system state derivative $\hat{\dot{\vec{x}}}$, which is finally integrated into the next system state $\vec{x}(t+h)$ by the \ac{ODE}-solver with time step size $h$. Also the \ac{FMU}'s input $\vec{u}$, output $\vec{y}$ or parameters (not shown) could be connected to the \acp{ANN}.}
	\label{fig:neuralfmume}
\end{figure}

We extend the concept of Neural\acp{ODE} by one or more \acp{FPM} in the form of \acp{FMU} to obtain a class of hybrid models, named \emph{NeuralFMUs} \cite{ThummererModelica:2021}. Using the example of a \ac{ME}-Neural\ac{FMU}, the \ac{ME}-\ac{FMU} replaces the \ac{ANN} of the Neural\ac{ODE}, because it calculates the system dynamics $\dot{\vec{x}}$ based on the current system state $\vec{x}$. To optimize the system state, an additional (state) \ac{ANN} can be placed \emph{before}, and to manipulate the system dynamics, an additional (derivative) \ac{ANN} can be placed \emph{after} the \ac{FMU}. This exemplified structure of a Neural\ac{FMU} is given in \reffig{neuralfmume}.
However, the concept of Neural\acp{FMU} itself is very generic and does not restrict the positions or number of \acp{FMU} inside the \acp{ANN} or limits which signals are manipulated by the \acp{ANN}. Inference of a Neural\ac{FMU} can be achieved by evaluating each of the considered blocks one after another. Whereas inference of the derivative \ac{ANN} (between \ac{FMU} and solver) is straight forward, because the system dynamics are passed as input to the \ac{ANN}, inference of the state \ac{ANN} needs additional attention. In case of an event inside the \ac{FMU}, the \ac{FMU} system state $\hat{\vec{x}}$ may be changed during event-handling. This new state must be propagated backwards through the state \ac{ANN} to calculate a new system state $\vec{x}$ for the \ac{ODE} solver to reinitialize the numerical integration at. Because \acp{ANN} are not invertible by default, the new state $\vec{x}$ must be determined by solving an optimization problem. In case of a state event, the required accuracy for the optimization solution is high, because solving for a state that is slightly before the event instance will trigger the event again. To prevent this, the optimization objective can be defined not only on hitting the \ac{FMU} state $\hat{\vec{x}}$, but also on the change of the sign of the corresponding event indicator. This enhanced objective promotes finding a state that lays after the event instance in time. In case of time events, high accuracy for the optimization result is desirable but not required.

For a more detailed view on the concept of Neural\acp{FMU} and the technical training process, see \cite{ThummererModelica:2021}. In the following, only \ac{ME}-Neural\acp{FMU} are considered and identified by the short term \emph{NeuralFMU}.

\subsection{Software}
Combining and training physical and data-driven models inside a single industry tool is currently not possible, therefore, it is required to transfer \acp{FPM} to a more suitable environment. After the export from modeling environment and import into the \ac{ML} environment, the first-principle is extended to a hybrid model. Finally after succeeded training, it is necessary to re-import the hybrid model back into the original modeling environment, for further modeling or the setup of larger system co-simulations. For the considered importing and exporting between environments, an industry typical model exchange format is needed. Because the \ac{FMI} is an open standard and widely used in industry as well as in research applications, it is a suitable candidate for this aim. On the side of modeling environments, \ac{FMI} is already implemented in many common tools, but a software interface that integrates \ac{FMI} into the \ac{ML} environment is still needed (s. \refsec{fmijl}). 

\subsubsection{Julia Programming Language}
In this section, it is shortly explained why the authors picked the Julia programming language (from here on simply referred to as \emph{Julia}) for the presented task as \ac{ML} environment. Julia is a dynamic typing language developing since 2009 and first published in 2012 \cite{Bezanson:2012}, with the aim to provide fast numerical computations in a platform-independent, high level programming language \cite{Bezanson:2015}. The language and interpreter was originally invented at the \emph{Massachusetts Institute of Technology}, but till today many other universities and research facilities have joined the development of language expansions, which mirrors in many contributions from different countries and even in its own conference, the \emph{JuliaCon}\footnote{\url{http://www.juliacon.org}}. Besides many great libraries in the field of scientific machine learning, there are multiple libraries for \ac{AD} like e.g. \emph{ForwardDiff.jl}\footnote{\url{https://github.com/JuliaDiff/ForwardDiff.jl}}\cite{Revels:2016} and \emph{Zygote.jl}\footnote{\url{https://github.com/FluxML/Zygote.jl}}\cite{Innes:2018}. Further, different modeling related libraries are available, like for example \libmodia{}\footnote{\urlmodia{}} \cite{Elmqvist:2018}, which allows for object-orientated white-box modeling of mechanical and electrical systems, syntactically similar to \emph{Modelica}\textsuperscript{\textregistered}, in Julia. 

\subsubsection{\ac{FMI} in Julia: \libfmi{}}\label{sec:fmijl}
The Julia library \libfmi{} provides high level commands to unzip, allocate, parameterize and simulate entire \acp{FMU}, as well as plotting the solution and parsing model meta data from the model description. Because \ac{FMI} has already three released specification versions and is under ongoing development, one major goal of \libfmi{} is to provide the ability to simulate different version \acp{FMU} with the same user front-end. To satisfy users who prefer close-to-specification programming, as well as users that are new to the topic and favor a smaller but more high level command set, we provide high level Julia commands, but also the possibility to use the more low level commands specified in the \ac{FMI} standards \cite{FMI:2020, FMI:2022}. The library and its feature set are constantly growing.

\subsubsection{Neural\acp{FMU} in Julia: \libfmiflux{}}
The open-source library \libfmiflux{} extends \libfmi{} and allows for the fusion of a \ac{FMU} and an \ac{ANN}. As in many other machine learning frameworks, a deep \ac{ANN} in Julia using \libflux{}\footnote{\urlflux{}} is configured by chaining multiple neural layers together. Probably the most intuitive way of integrating a \ac{FMU} into this topology, is to simply handle the \ac{FMU} as a network layer. In general, \libfmiflux{} does not make restrictions to:
\begin{itemize}
	\item which \ac{FMU} signals can be used as layer inputs and outputs. It is possible to use any variable that can be set via \mycode{fmi2SetReal} or \mycode{fmi2SetContinuousStates} as input and any variable that can be obtained by \mycode{fmi2GetReal} or \mycode{fmi2GetDerivatives} as output;
	\item where to place \acp{FMU} inside the \ac{ANN} topology, as long as all signals are traceable via \ac{AD} (no signal cuts).
\end{itemize}
Dependent on the \ac{FMU} type, \ac{ME}, \ac{CS} or \ac{SE}, different setups for NeuralFMUs should be considered. In this article, only \ac{ME}-Neural\acp{FMU} are highlighted.

\subsection{Workflow}
On the foundation of Julia, \ac{FMI}, \libfmi{} and \libfmiflux{}, we suggest the following workflow for designing custom NeuralFMUs:
\begin{figure}[h!]
	\includegraphics[width=\textwidth]{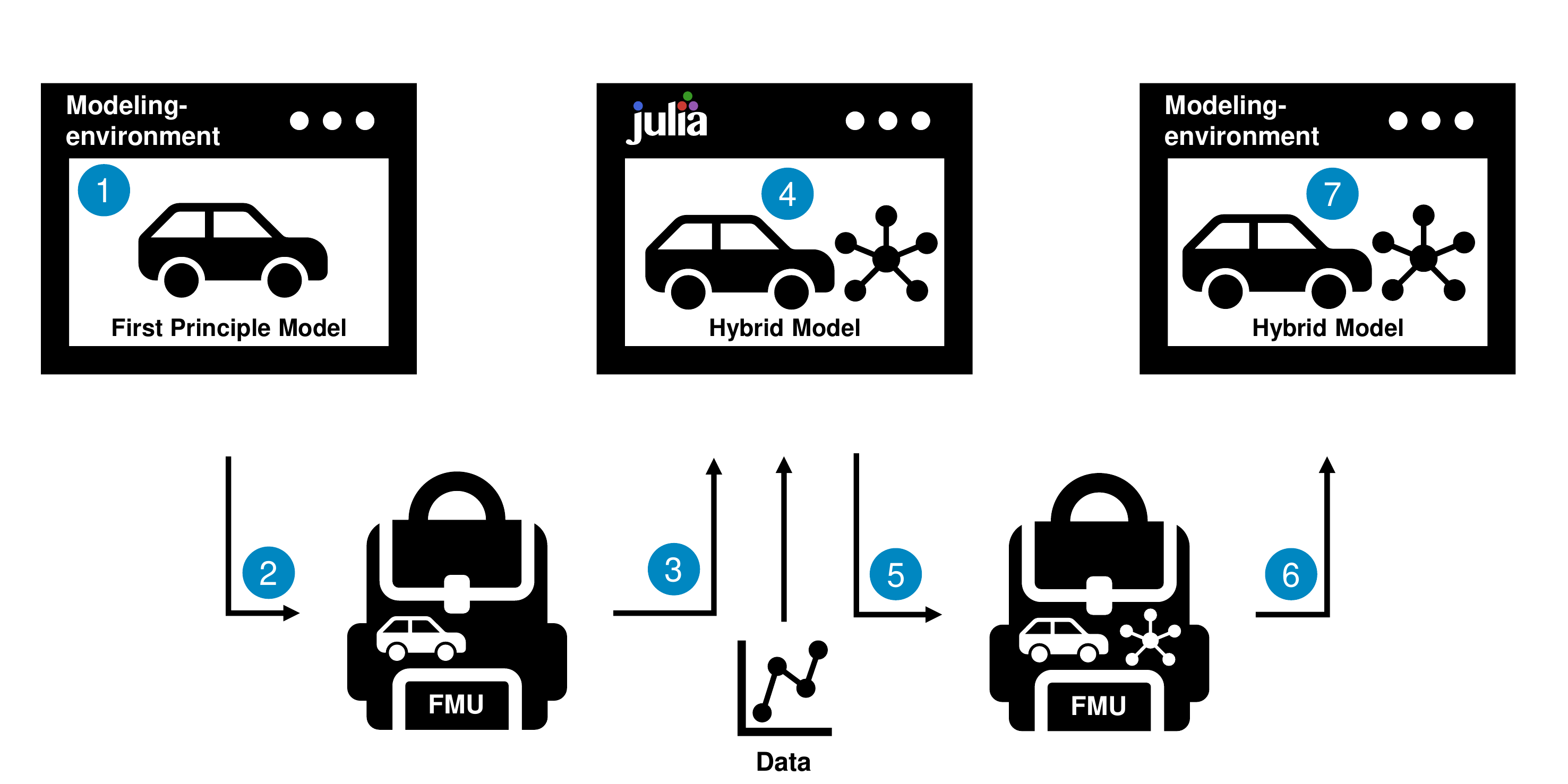}
	\caption{Workflow of the presented hybrid modeling application using a modeling tool supporting \ac{FMI} (export and import) and \libfmi{} together with \libfmiflux{}.\label{fig:workflow}}
\end{figure}   

The presented development process of a NeuralODE/FMU in \reffig{workflow} covers the following steps:
\begin{enumerate}
	\item The \ac{FPM} is designed by a domain expert inside of a familiar modeling tool supporting \ac{FMI} (export and import).
	\item After modeling, the \ac{FPM} is exported as \ac{FMU}.
	\item The \ac{FPM}-\ac{FMU} is imported into Julia using \libfmi{}.
	\item The \ac{FPM} is extended to a hybrid model and trained on data, for example of a real system or a high-resolution and high-fidelity simulation, using \libfmiflux{}.
	\item The trained hybrid model is exported as \ac{FMU} using \libfmi{}.
	\item The hybrid model \ac{FMU} is imported into the original modeling environment.
	\item The improved hybrid model \ac{FMU} can further be used in different modeling and simulation tools (including the original tool the \ac{FPM} was exported from).
\end{enumerate}

\subsection{Data pre- and post-processing}\label{sec:prepost}
In many \ac{ML} applications, pre- and post-processing are not only instrumental, but necessary. For training conventional \acp{ANN}, pre-processing of training data can often be performed for one single time, before batching and the actual training. For Neural\acp{FMU}, the \acp{FMU} may generate outputs within a range that is excessively saturated by the activation functions inside the \ac{ANN}. Further, the \acp{FMU} may expect inputs within a range not generatable by the \ac{ANN}, because of the limited output of the activation functions. Therefore, all signals must be processed at the interfaces between the \acp{FMU} and \acp{ANN}.
If no expert knowledge of the data range of the \ac{FMU} inputs and outputs is available, a good starting point can be to scale and shift data into a standard normal distribution. Because the \ac{FMU} output and input may shrink or grow during training because of new state exploration by the changed dynamics, scaling and shifting parameters should be parts of the optimization parameters during training. See \refsec{usedtopology} for a visual example topology using data pre- and post-processing around an \ac{ANN}.

\subsection{Initialization (Pre-training)}\label{sec:initialization}
Obtaining a trainable (solvable) NeuralFMU is not trivial. Using larger or complex \acp{FMU} together with randomly initialized \acp{ANN} often leads to instable and/or stiff ODE-system. Whereas starting the training process with an unnecessary stiff NeuralFMU (more stiff than the final solution) leads to long training times, an instable system might not be trainable at all. Without further investigation, selecting random initialized \acp{ANN} as part of NeuralFMUs often lead to hardly trainable systems in different use cases like e.g. a controlled EC-Motor \ac{HiL}-simulation, a thermodynamic cabin simulation or in modeling the human cardiovascular system \cite{ThummererICMSquare:2021}. Therefore, we suggest three different initialization strategies for \acp{ANN} inside of NeuralFMUs: \ac{NIPT}, \ac{CCPT} and the introduction of a \ac{FPM}/\ac{ANN} gates topology. A major advantage of all initialization modes is, that sensitivities during the initialization process don't need to be propagated through the \ac{ODE} solver (the actual \ac{ODE} is not solved), therefore, the computational effort is much less, compared to the actual training described in \refsec{training}.

For a better understanding, initialization strategies are not exemplified at a NeuralFMU as in \reffig{neuralfmume}, but a suitable Neural\ac{FMU} topology for this use case, which includes only one \ac{FMU}, one \ac{ANN} (derivative) and the numerical solver. The concepts can be easily modified to fit other topologies.

\subsubsection{\acl{NIPT} (\acs{NIPT})}
\begin{figure}[h!]
	\centering
	\includegraphics[height=3.5cm]{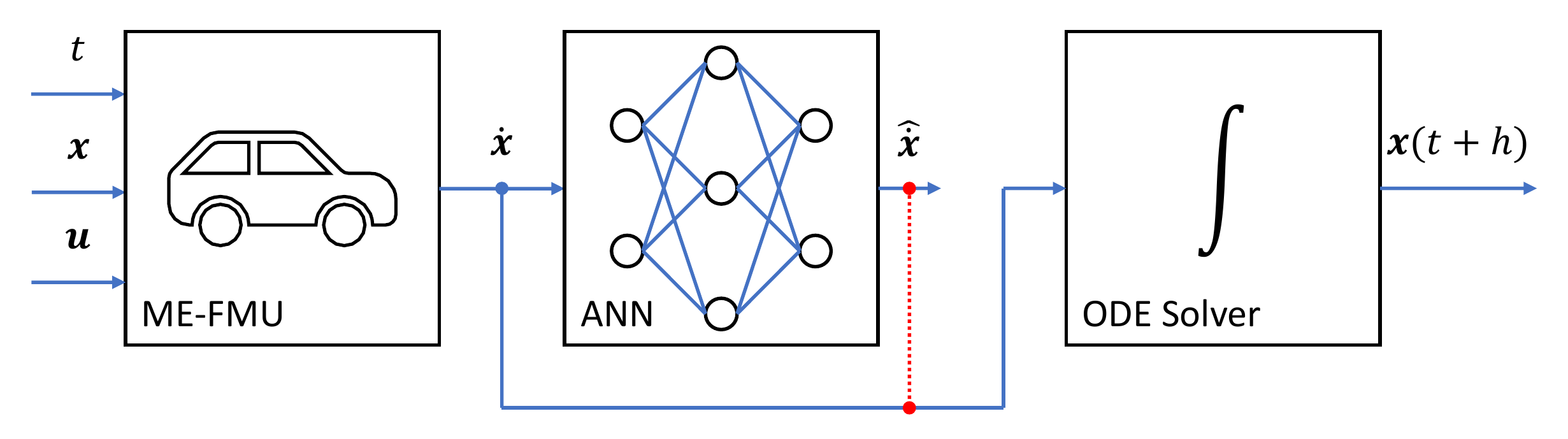} 
	\caption{For \ac{NIPT}, a single reference simulation with unchanged $\dot{\vec{x}}$ is performed to calculate the \ac{ANN} output $\hat{\dot{\vec{x}}}(t)$ for every system state $\vec{x}(t)$. Based on that, the actual training can be performed. As soon as the training goal $\dot{\vec{x}} \approx \hat{\dot{\vec{x}}}$ (red-dotted) is reached, the NeuralFMU dynamics equals the dynamics of the \ac{FMU} itself, which results in the same solution for both. In this case, the \ac{ANN} behaves \emph{neutral}.}
	\label{fig:neutralinit}
\end{figure}

If the system state derivative is not known, cannot be measured and/or can hardly be approximated, \ac{NIPT} of the \acp{ANN} can deliver a good initialization for the \ac{ANN} parameters for the later training. Similar to auto-encoder networks, the aim is to train the \ac{ANN} so, that the output equals the input for a set of training data, see \reffig{neutralinit}. Different from auto-encoders, the hidden network layers don't need to narrow in width. The training result is, that the solution of the initialized NeuralFMU converges against the solution of the \ac{FMU} itself - or if multiple \acp{FMU} are present - the solution of the chained \ac{FMU}-system without \acp{ANN}. If the \ac{FMU} solution is already close to the target solution (the term \emph{close} strongly depends on the system constitution), this might be a suitable initialization method. Only for training data aquisition, it is necessary to perform a single forward simulation. For the actual training, solving the \ac{ODE} system is not required.


\subsubsection{\acl{CCPT} (\acs{CCPT})}
\begin{figure}[h!]
	\centering
	\includegraphics[height=3.5cm]{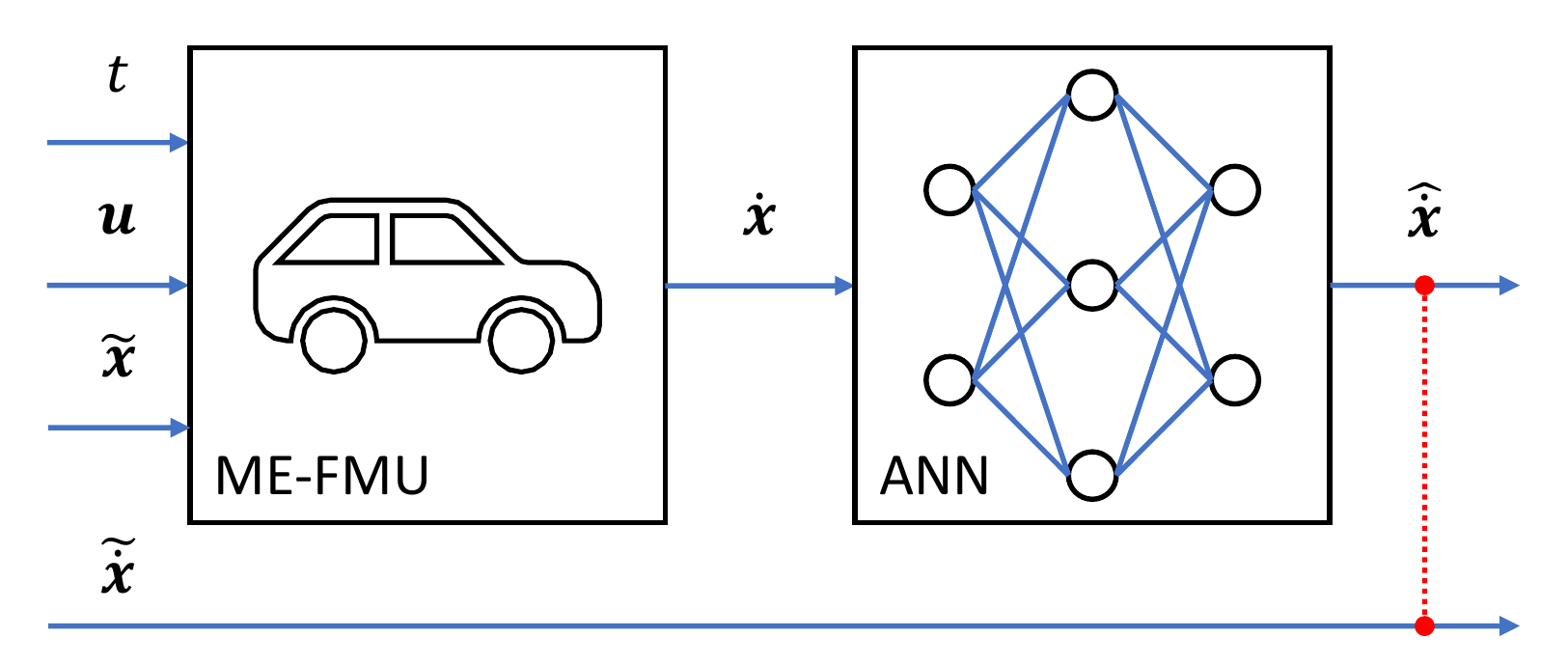} 
	\caption{For \ac{CCPT}, no reference simulation is performed. For a given state trajectory $\tilde{\vec{x}}(t)$ (e.g. from data), every state is propagated through the \ac{FMU} and the \ac{ANN}. As soon as the training goal $\hat{\dot{\vec{x}}} \approx \tilde{\dot{\vec{x}}}$ (red-dotted) is reached, the NeuralFMU dynamics equals the target dynamics $\tilde{\dot{\vec{x}}}$ (e.g. from data). As a result, the later NeuralFMU solution matches the given state trajectory for a perfect known $\tilde{\dot{\vec{x}}}$. The target dynamics $\tilde{\dot{\vec{x}}}$ may be estimated by deriving and filtering the given system state $\tilde{\vec{x}}(t)$ or might be known from measurements.}
	\label{fig:collocation}
\end{figure}

Similar to the collocation training of Neural\acp{ODE} \cite{Roesch:2021}, collocation training can be performed for Neural\acp{FMU}, too. The \ac{CCPT} is similar to \ac{NIPT}, the major difference is the training goal, see \reffig{collocation}. Whereas \ac{NIPT} focuses on propagating the derivatives unchanged through the \ac{ANN}, \ac{CCPT} aims on hitting the derivatives of the \ac{ODE}-solution, so that after integration (solving), the target solution can be obtained.

This method needs knowledge of the entire system state trajectory $\tilde{\vec{x}}(t)$ as well as the (at least approximated) state derivative $\tilde{\dot{\vec{x}}}(t)$. In general, only a part of the system state and/or derivative of a real system can be measured. Different methods allow for estimating the unknown states like e.g. Kalman-filter \cite{Kalman:1960}. To converge against the target solution, \ac{CCPT} needs high-quality data of the system state and derivative. Derivatives could be approximated by finite differences or filters, see \cite{Roesch:2021} for an overview. Note that approximating the derivatives may decrease the quality of the pre-training process.

\ac{CCPT} is only usable, if the states of the training objective matching the state derivatives manipulated by the \ac{ANN}. Whereas this is often the case in academic examples, in real applications it is not, which is further exemplified at the \ac{VLDM} in \refsec{usedtopology}.
\\
To summarize, \ac{NIPT} doesn't require the entire state, but converges only against the \ac{FMU} solution. \ac{CCPT} on the other hand, converges against a given target solution, but requires a known target solution and derivative.


\subsubsection{\ac{FPM}/\ac{ANN} gates}\label{sec:gates}
\begin{figure}[h!]
	\centering
	\includegraphics[height=4.5cm]{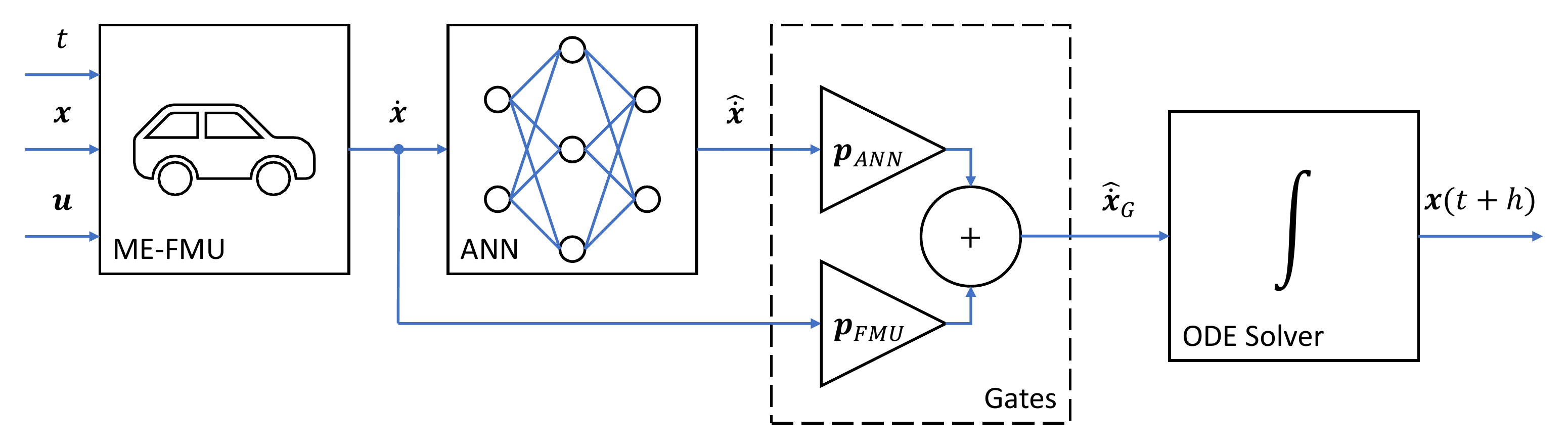} 
	\caption{\ac{ME}-NeuralFMU with \ac{FPM}/\ac{ANN} gates: The \ac{ME}-\ac{FMU} receives the current time $t$ and state $\vec{x}$ from the numerical ODE solver together with an external input $\vec{u}$ and computes the corresponding state derivative $\dot{\vec{x}}$. Two gates $\vec{p}_{ANN}$ and $\vec{p}_{FMU}$ scale how much dynamic changes by the \ac{ANN} and the \ac{FMU} are introduced to the derivative vector $\hat{\dot{\vec{x}}}_G$. Finally, the derivative vector $\hat{\dot{\vec{x}}}_G$ is passed to the \ac{ODE} solver with (adaptive) step size $h$ and integrated to the next system state $\vec{x}(t+h)$.}
	\label{fig:gates}
\end{figure}

The challenge of finding a good initialization by a foregoing pre-training procedure can be bypassed by introducing a slightly modified topology, that literally introduces a bypass around the \ac{ANN}, see \reffig{gates}. 

As in \reffig{gates}, the system state derivative $\hat{\dot{\vec{x}}}_G$ is defined as follows, where $\circ$ stands for the Hadamard-product:
\begin{equation}
\hat{\dot{\vec{x}}}_G = \vec{p}_{ANN} \circ \hat{\dot{\vec{x}}} + \vec{p}_{FMU} \circ \dot{\vec{x}}
\end{equation}

On the one hand, for the case $\vec{p}_{ANN}=\vec{0}$ and $\vec{p}_{FMU}=\vec{1}$, the resulting simulation trajectory is just the simulation trajectory of the original \ac{FMU}, independent of the \ac{ANN} parameters. In this way, the Neural\ac{FMU} can be initialized without a special initialization routine, while also being capable of manipulating the system dynamics if the parameter $\vec{p}_{ANN}$ is changed to a non-zero value. On the other hand, for the case $\vec{p}_{ANN}=\vec{1}$ and $p_{FMU}=\vec{0}$, only the \ac{ANN} affects the state dynamics and the original \ac{FMU} dynamics are used only as input for the \ac{ANN}. The parameters $\vec{p}_{ANN}$ and $\vec{p}_{FMU}$ can be optimized along the other training parameters, or dependent on the use case, with a static or dynamic decay/increase. As a final note, \ac{CCPT} and the \ac{FPM}/\ac{ANN} gates topology doesn't exclude each other and can be used together on a Neural\ac{FMU} initialization.

\subsection{Batching \& Training}\label{sec:training}
The training is not performed on the entire data trajectory at one time (e.g. the used \traincycle{} is $1001.22\,s$ long). Instead the trajectory is batched. The major challenge at this point is that for a given batch element start time $t \neq t_{0}$, often only the continuous part of the model state vector $\vec{x}_c(t)$ of the \ac{ME}-\ac{FMU} is known. In general, the discrete part $\vec{x}_d(t)$ is unknown. As a result, if not explicitly given, a suitable discrete state is determined during the initialization procedure inside the \ac{FMU}, but it is not guaranteed that this state matches the data and/or expectations. This circumstance also applies to the determination of the initial value of the discrete states $\vec{t_0}_d$, but measurements are often started in a stationary state, where a good understanding of the correct discrete states is given, even if they are not explicitly parts of the data measurements.
\\
As a consequence, training cannot be initialized at an arbitrary element of the batch (time instant) because of the unknown discrete states,  which might be initialized unexpected if ignored. Estimating the discrete system state on basis of data is not trivial and may need significant expert knowledge about the model itself. So, a straight-forward strategy to handle this is to simulate all batches in the correct order without resetting the \ac{FMU} between batch elements. Although this does not guarantee the correct discrete state when switching from one batch element to the next during training, the discrete solution converges against the target together with the continuous solution.
\\
Another option is to simulate the entire trajectory for one single time and making memory copies of the entire \ac{FMU} state using e.g. \mycode{fmi2GetFMUstate} (in \ac{FMI} 2.0) at the very beginning of every batch element. This allows for random batches, which might improve the training success and speed. Because the feature required to save and load the \ac{FMU} memory footprint is optional in \ac{FMI} and thus often not implemented, this strategy is not further highlighted at this point, but can be implemented in a straight-forward manner.

Besides the ones mentioned, many techniques for NeuralODEs can be adapted to improve the training process in terms of stability and convergence, like for example \emph{multiple shooting} as in \cite{Turan:2022}, because NeuralFMUs are a subclass of NeuralODEs.

\section{Results}\label{sec:results}
The considered method is validated at the following application: Based on the introduced \ac{VLDM}, a hybrid model is deployed leading to a significant better consumption prediction than the original \ac{FPM}. Even if the \ac{FPM} already considers multiple friction effects, it is assumed that the prediction error is the result of a wrongly parameterized or an additional, non-modeled friction effect. The goal is to inject an additional vehicle acceleration to force the driver controller to perform a higher engine torque, and thus to increase the vehicle energy consumption.
\subsection{Example Model: \acl{VLDM} (\acs{VLDM})}\label{sec:VLDM}
\begin{figure}[h!]
	\includegraphics[width=\textwidth]{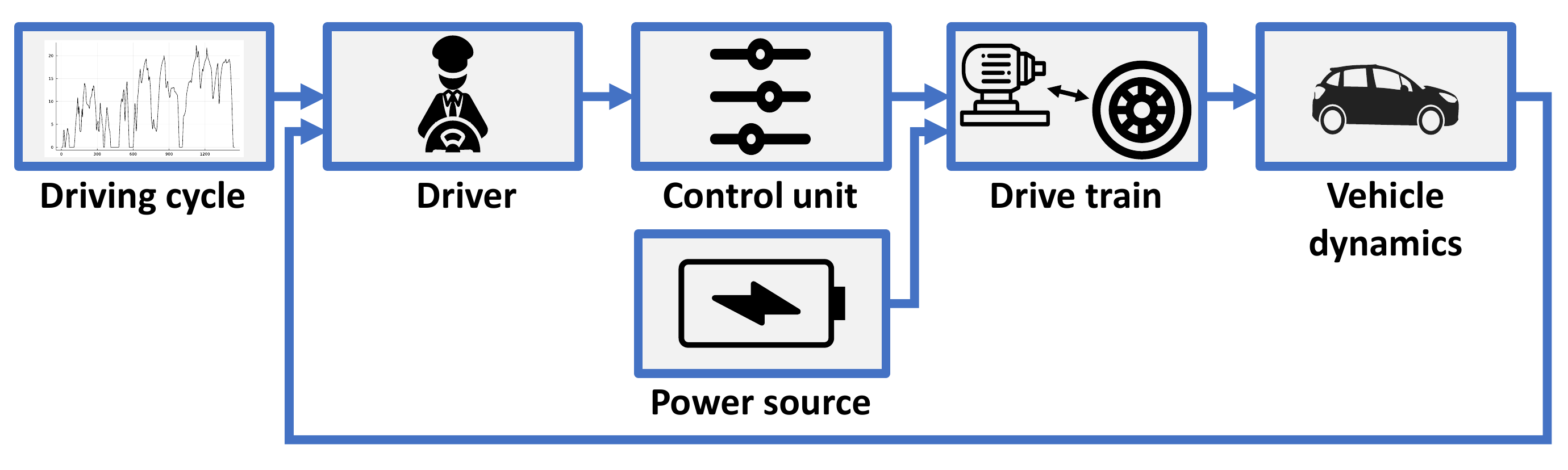}
	\caption{Topology of the \ac{VLDM} (own representation adapted from \cite{Danquah:2019}).}
	\label{fig:model}
\end{figure}

In this section, the used \ac{FMU} model is introduced: The model represents the longitudinal dynamics of an electric \emph{Smart EQ fortwo} and is an adaption from a model of the \emph{Technical University of Munich} \cite{Danquah:2019}. In automotive applications, longitudinal dynamics models are often used to simulate energy consumption, thus only related effects are represented in the model. The original model was created in MATLAB\textsuperscript{\textregistered}/Simulink\textsuperscript{\textregistered}, replicated analogously in the modeling language Modelica\textsuperscript{\textregistered} and exported as \ac{FMU}. The following \reffig{model} shows the topology of the simulation model. The full vehicle model is modular and consists of six core components \cite{Guzzella:2013}.

The top-layer components of the vehicle are the \textit{Driving cycle}, the \textit{Driver}, the \textit{Control unit}, the \textit{Power source}, the \textit{Drive train} and the \textit{Vehicle dynamics} subsystem. The target vehicle speed, read from a driving cycle tabular in the \textit{Driving cycle} subsystem, is forwarded to the \emph{Driver}. The \textit{Driver} consists of two PI-controllers for the acceleration and brake pedal, controlling the vehicle to follow the given driving cycle. The block \textit{Control unit} calculates the desired torque from the pedal position. The required power is provided by the \textit{Power source} component, modeling the vehicle battery. In the \textit{Drive train} block, the sources of acceleration and braking torques are implemented. This component contains the electric motor, the power electronics, the transmission and the tire model. In the component \textit{Vehicle dynamics}, the rolling-, air- and slope-resistance are implemented. The resulting force and torque determine the acceleration of the vehicle, and after numerical integration the speed and position \cite{Danquah:2019}. Finally, the vehicle speed is fed back into the \textit{Driver} subsystem and closes the controller loop. Together with the model itself, multiple measurements from an automotive driving test rig with different driving cycles like the \ac{CADC}, \ac{NEDC} and \ac{WLTC} where published\footnote{\url{https://github.com/TUMFTM/Component_Library_for_Full_Vehicle_Simulations}}. We use the driving cycles \traincycle{} and \testcycle{} for the presented experiment.
\\
The simulation model counts six continuous states $\vec{x}_c$ in total:
\begin{itemize}
	\item $x_1$ the PI-controller state (integrated error) for the throttle pedal (\textit{Driver});
	\item $x_2$ the PI-controller state (integrated error) for the brake pedal (\textit{Driver});
	\item $x_3$ the integrated driving cycle speed, the cycle position (\textit{Driving cycle});
	\item $x_4$ the vehicle position (\textit{Vehicle dynamics});
	\item $x_5$ the vehicle velocity (\textit{Vehicle dynamics});
	\item $x_6$ the cumulative consumption (energy).
\end{itemize}
In analogy, the six continuous state derivatives $\dot{\vec{x}}_c$ are:
\begin{itemize}
	\item $\dot{x}_1$ the PI-controller error/deviation for the throttle pedal (\textit{Driver});
	\item $\dot{x}_2$ the PI-controller error/deviation for the brake pedal (\textit{Driver});
	\item $\dot{x}_3$ the driving cycle speed (\textit{Driving cycle});
	\item $\dot{x}_4$ the vehicle velocity (\textit{Vehicle dynamics});
	\item $\dot{x}_5$ the vehicle acceleration (\textit{Vehicle dynamics});
	\item $\dot{x}_6$ the current consumption (power).
\end{itemize}
Note that this system features different challenging attributes, like:
\begin{itemize}
	\item   The system is highly discontinuous, meaning it has a significant amount of explicitly time-dependent events (100 events per second). This further limits the maximum time step size for the numerical solver and so worsens the simulation and training performance;
	\item   The simulation contains a closed-loop over multiple subsystems with two controllers running at $100\,Hz$ (the source of the high-frequent time-events);
	\item   The system contains a large amount of state-dependent events, triggered by 22 event-indicators;
	\item 	Measurements of the real system are not equidistant in time (even if it was saved this way, which introduces a typical measurement error);
	\item   Only a subpart of the system state vector is part of measurements, the remaining parts are estimated;
	\item 	Measurements are not exact and contain typical, sensor- and filter-specific errors (like noise and oscillation);
	\item   There is a hysteresis loop for the activation of the throttle and brake pedal, the PI-controller states are initialized at corresponding edges in the hysteresis loop;
	\item   The system is highly non-linear, e.g. multiple quantities are saturated;
	\item   Characteristic maps (data models) for the electric power, inverted electric power and the electric power losses are used.
\end{itemize}
Combining all these attributes results in a challenging \ac{FPM} for the considered hybrid modeling use case.

\subsection{Topology}\label{sec:usedtopology}
\begin{figure}[h!]
		\centering
		\includegraphics[height=4.5cm]{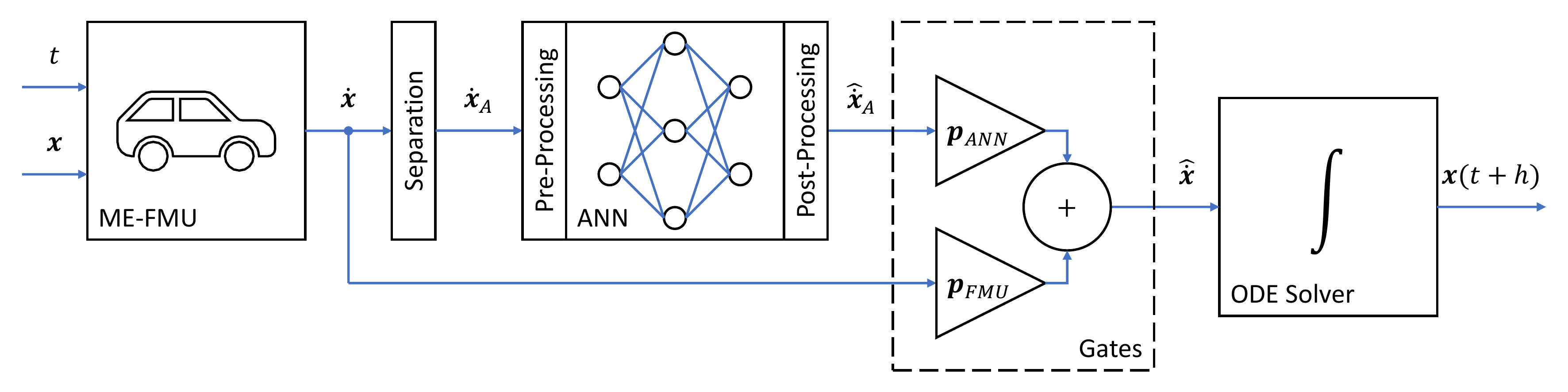} 
	\caption{Topology of the used \ac{ME}-NeuralFMU: The \ac{ME}-\ac{FMU} receives the current time $t$ and state $\vec{x}$ from the numerical ODE solver and computes the corresponding state derivative $\dot{\vec{x}}$. The state derivative is then trimmed to a subset of derivatives $\dot{\vec{x}}_A$. Before the \ac{ANN} transforms $\dot{\vec{x}}_A$, the signals are pre-processed to approximately fit a standard normal distribution. After the \ac{ANN} inference, the inverse transformation of the pre-process is applied in the post-processed and $\hat{\dot{\vec{x}}}_A$ is obtained. Two gates $\vec{p}_{ANN}$ and $\vec{p}_{FMU}$ scale how much $\hat{\dot{\vec{x}}}_A$ and $\dot{\vec{x}}$ are introduced to the final derivative vector $\hat{\dot{\vec{x}}}$. Finally, the changed derivative vector $\hat{\dot{\vec{x}}}$ is passed to the \ac{ODE} solver with (adaptive) step size $h$ and integrated to the next system state $\vec{x}(t+h)$. The used \ac{FMU} has no continuous inputs $\vec{u}$, the driving cycle is part of the model and depends only on the time $t$.}
	\label{fig:usedneuralfmu}
\end{figure}

Combining the orignal Neural\ac{FMU} topology (s. \reffig{neuralfmume}) with pre- and post-processing (s. \refsec{prepost}) and \ac{FPM}/\ac{ANN} gates (s. \refsec{gates}) results in the topology as shown in \reffig{usedneuralfmu}, which is used for training the hybrid model in the considered use case.
In general, understanding at least some aspects of the physical effect, which is to be modeled by the \acp{ANN}, is a great advantage. Basically, any value of the \ac{FMU} that is accessible via the \ac{FMI} can be used as input for the \acp{ANN}: The system states and -derivatives, system inputs as well as any other system variable (or output) that depends on the system state, input and/or time. This allows for a wide variety of NeuralFMU topologies. However, using all available variables in the interface to the \acp{ANN} can result in sub-optimal training performance, because more variable sensitivities need to be determined and signals without physical dependency could be misinterpreted as dependent. This motivates the use of a clever, minimal subset of the available \ac{FMU} variables. Often, state and state derivatives are good choices for variables to feed into the \ac{ANN}, because from a mathematical point of view, the system state holds all relevant information in a minimal representation. Nevertheless, adding additional variables may be productive, even if the encapsulated system information is redundant, if the correlations between these variables and the training objective is easier to fit for an \ac{ANN} compared to the correlation with system states and/or derivatives. 
\\
For the considered use case, a friction-effect is learned. Conventional mechanical friction models, like viscous damping or slip-stick-friction, often depend on the physical body's translational or rotational velocity. Therefore, especially the vehicle speed should be considered. Further, the current vehicle acceleration and power is also given as inputs to the \ac{ANN}. The training objective is to match the cumulative consumption from training data by manipulating the vehicle acceleration. Here, \ac{CCPT} can't be used, because the \ac{CCPT} objective would be to fit the cumulative consumption derivative, so the current consumption, but this value is not affected by the \ac{ANN}.

To summarize, the following variables (compare to \reffig{usedneuralfmu}) are used:
\begin{itemize}
	\item $\dot{\vec{x}}_A = \{\dot{x}_4, \dot{x}_5, \dot{x}_6\}$ corresponds to the vehicle speed, acceleration and power (current consumption). These are the inputs for the \ac{ANN};
	\item $\hat{\dot{\vec{x}}}_A = \{0, 0, 0, 0, \hat{\dot{x}}_5, 0\}$ corresponds to the estimated vehicle acceleration by the \ac{ANN}. This is the output of the \ac{ANN} (technically, it is the only output, because the other five dynamics are assumed always zero and are neglected);
	\item $\vec{p}_{ANN} = \{0, 0, 0, 0, p_1, 0\}$, only the influence of the vehicle acceleration from the \ac{ANN} can be controlled via $p_1$ (this is the only \ac{ANN} output);
	\item $\vec{p}_{FMU} = \{1, 1, 1, 1, p_2, 1\}$, only the influence of the vehicle acceleration from the \ac{FMU} can be controlled via $p_2$ (all other derivatives contribute by 100\%).
\end{itemize} 
Please note that if the considered effect could depend on the system states, states could also be passed as input to the \ac{ANN}. Because the hybrid model reuses the \ac{FPM} and therefore the \ac{ANN} only needs to approximate the unmodeled physical effect, a very light-weight net layout is sufficient, as shown in \reftab{layout}. This results in a fast training because of the small number of parameters. 
\begin{table}[h!] 
	\caption{ANN layout and parameters of the used topology.\label{tab:layout}}
	\newcolumntype{C}{>{\centering\arraybackslash}X}
	\begin{tabularx}{\textwidth}{CCCrrrr}
		\toprule
		Index & Type & Activation & Inputs & Outputs & Bias & Parameters\\
		\midrule
		1 & Pre-process& none  & 3 & 3 & 0 & 6 \\
		2 & Dense& tanh  & 3 & 32 & 32 & 128 \\
		3 & Dense& tanh  & 32 & 1 & 1 & 33 \\
		4 & Post-process & none  & 1 & 1 & 0 & 0\\
		5 & Gates & none  & 1 & 1 & 0 & 2 \\
		\bottomrule
		~ &~ &~ &~&~ & ~ & Sum: 169
	\end{tabularx}
\end{table}

\subsection{Consumption prediction}
\subsubsection{Training}
After training on a batch with a batch element length of \batchlen{}, resulting in \numbatches{} batch elements, and \numepochs{} epochs on the \traincycle{}, the NeuralFMU is able to outperform the \ac{FPM} on training data, see \reffigs{restrain}{restrain10}{ertrain}. The following figures show the predicted cumulative consumption over time of the original \ac{FPM}, compared to the trained NeuralFMU. Training is not converged at this point. The cost function is implemented as ordinary \ac{MSE} between data and predicted cumulative consumption. For parameter optimization \emph{Adam} \cite{Kingma:2014} together with an exponential decay\footnote{Initial step size: $1e-3$, Decay (new step size multiplier): $0.95$ every step, Min. step size: $1e-5$.} is used. The training is performed single-core on CPU\footnote{Intel\textregistered{} Core\textsuperscript{TM} i7-8565U on Windows 10 Enterprise 20H2} and takes \trainlength{}. During interpretation of results, please note the small amount of data used for training: A single driving cycle measurement trajectory (mean over two real experiments).
\begin{figure}[h!]
	\centering
	\includegraphics[width=12cm]{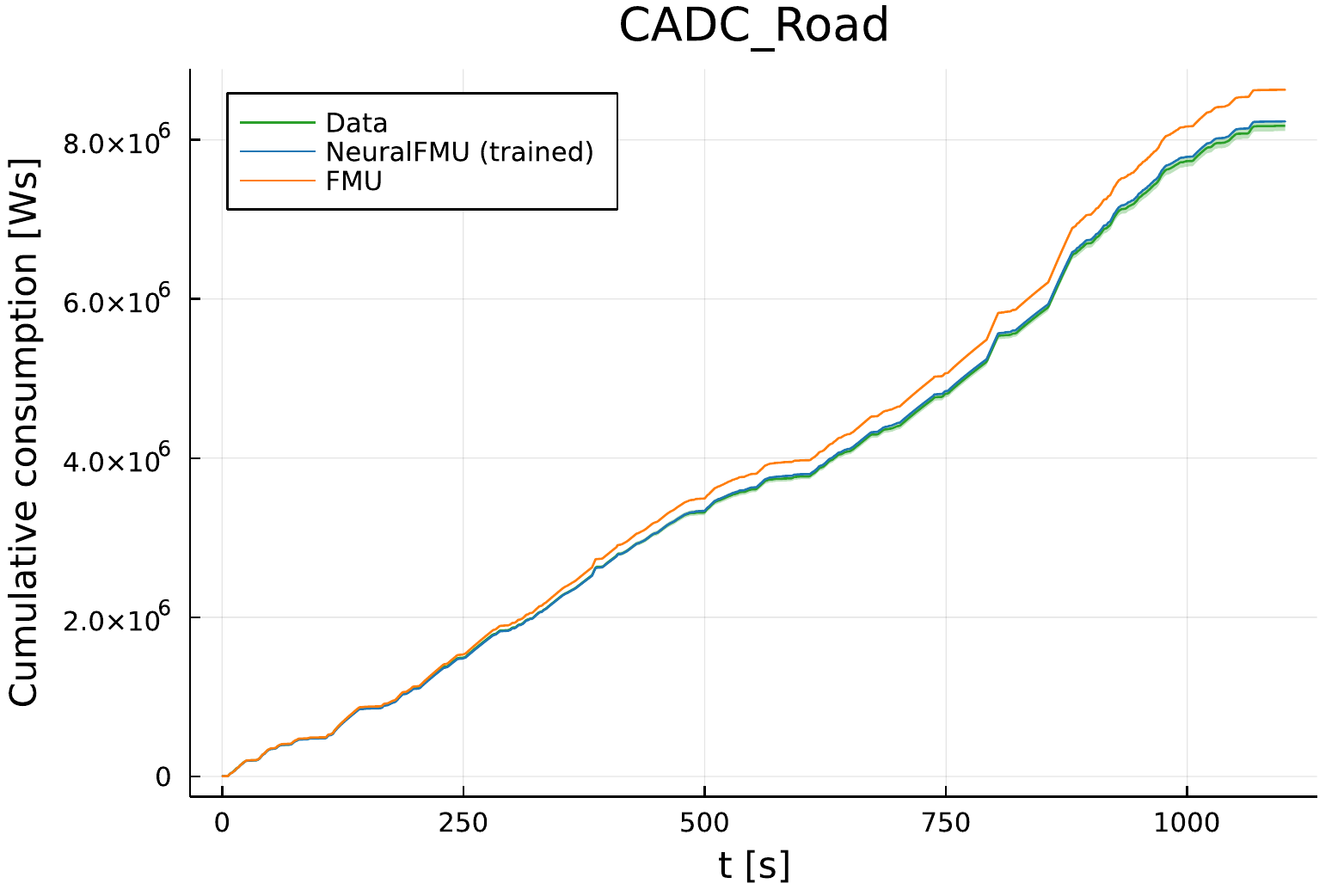}
	\caption{Cumulative consumption prediction on the \traincycle{}, that is part of training data. The NeuralFMU (blue) lays almost on the training data mean (green) and inside the data uncertainty region (green, translucent). On the other hand, the simulation results of the original \ac{FPM} as \ac{FMU} (orange) slowly drifts out the data uncertainty region, resulting in a relative large error at the simulation stop time compared to the NeuralFMU.}
	\label{fig:restrain}
\end{figure} 
\begin{figure}[h!]
	\centering
	\includegraphics[width=12cm]{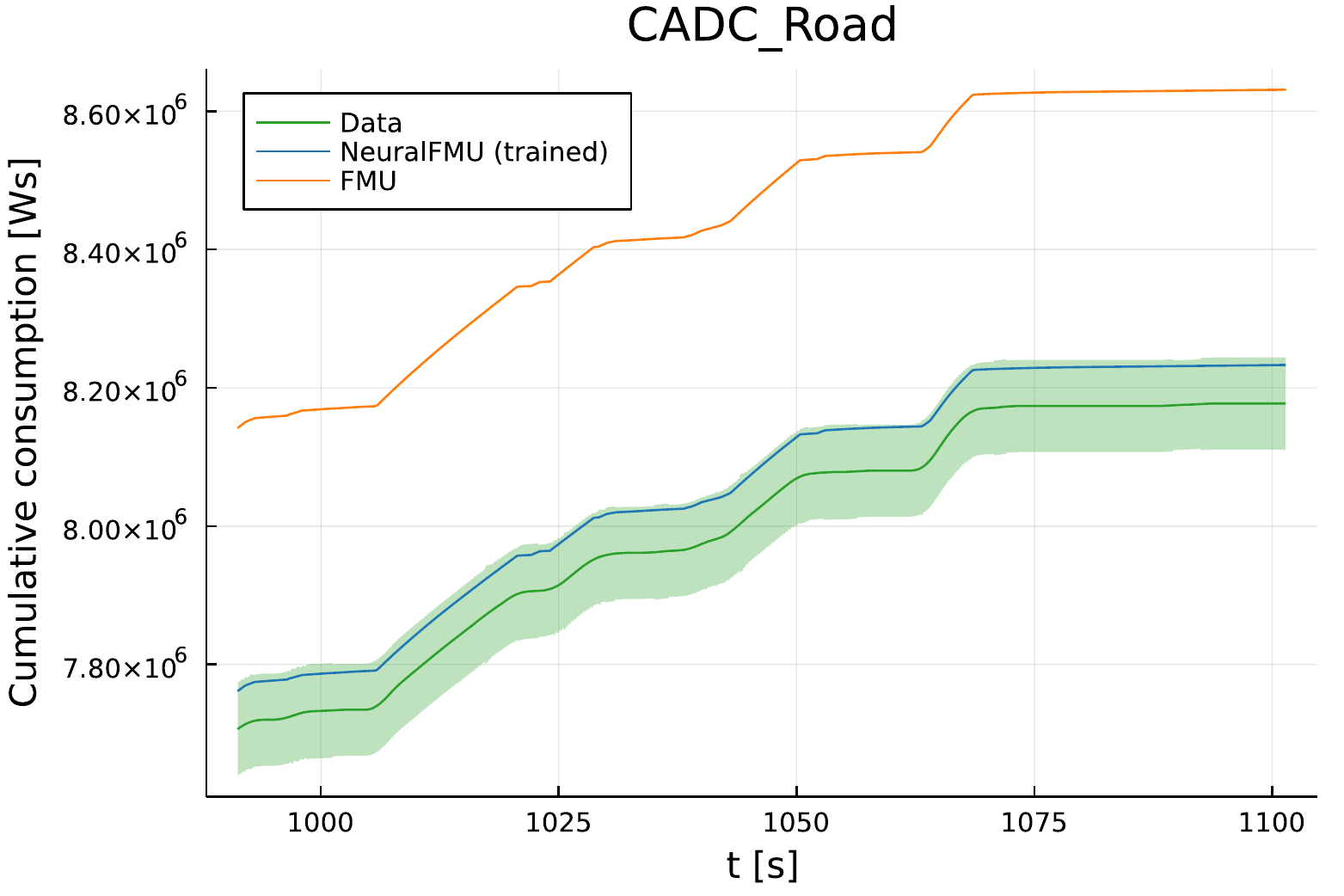}
	\caption{Deviation on the last $10\,\%$ of the cumulative consumption prediction on the \traincycle{}, that is part of training data. The final consumption prediction accuracy of the NeuralFMU (blue) significantly increases compared to the original \ac{FMU} (orange), lays inside the measurement uncertainty (green, translucent) and close to the data mean (green). The original \ac{FPM} prediction lays completely outside of the measurement uncertainty.}
	\label{fig:restrain10}
\end{figure} 
\begin{figure}[h!]
	\centering
	\includegraphics[width=12cm]{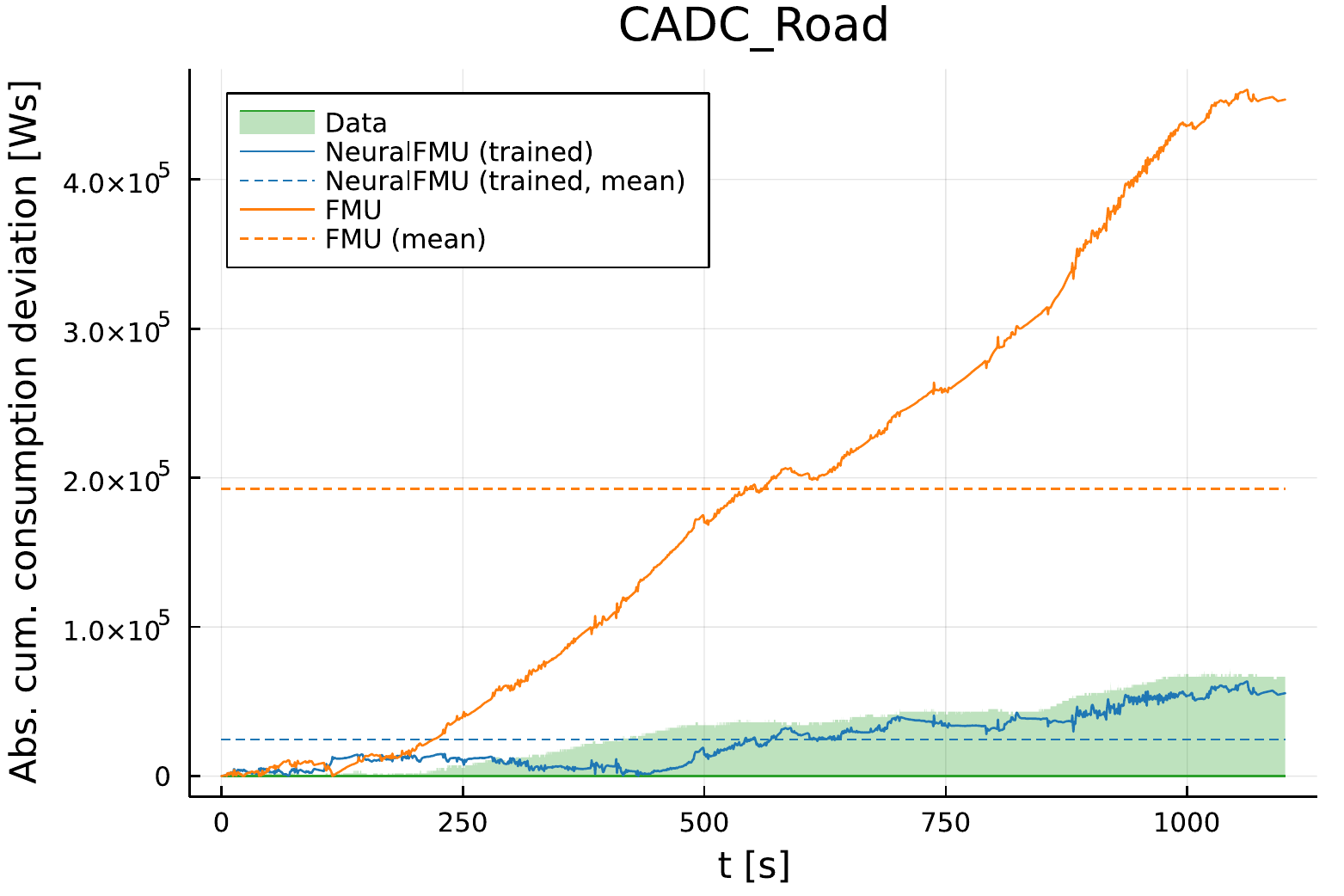}
	\caption{Deviation (absolute error) of the consumption prediction between data and the NeuralFMU (blue) compared to the original \ac{FPM} as \ac{FMU} (orange) on the \traincycle{}, that was part of training data. After $\approx 300\,s$, the NeuralFMU solution lays inside the data uncertainty region (green, translucent) and outperforms the \ac{FPM} in terms of prediction accuracy. Further, the NeuralFMU error is at any time step significantly smaller than the mean error of the FMU (orange, dashed).}
	\label{fig:ertrain}
\end{figure}

\clearpage
\subsubsection{Testing (Validation)}
\begin{figure}[h!]
	\centering
	\includegraphics[width=12cm]{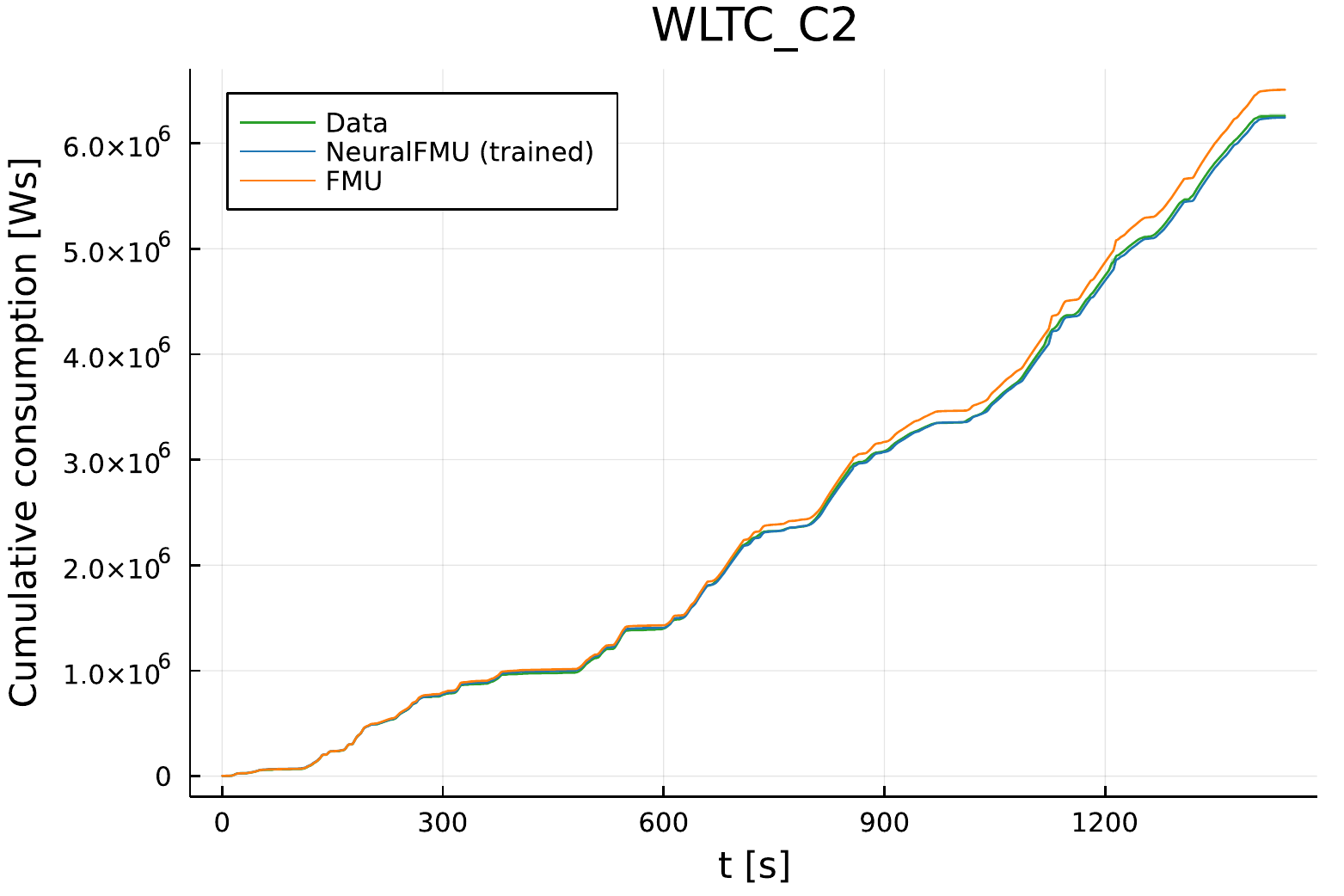}
	\caption{Comparison of the cumulative consumption prediction, using the \testcycle{}, that is not part of the NeuralFMU training data. As for training data, the NeuralFMU prediction (blue) is closer to the measurement data mean (green), compared to the original \ac{FPM} prediction (orange).}
	\label{fig:restest}
\end{figure}

After training, the Neural\ac{FMU} is validated against unknown data by simulating another driving cycle, the \testcycle{}, that is not known from training. Results and explanations can be seen in \reffigs{restest}{restest10}{ertest}.
\begin{figure}[h!]
	\centering
	\includegraphics[width=12cm]{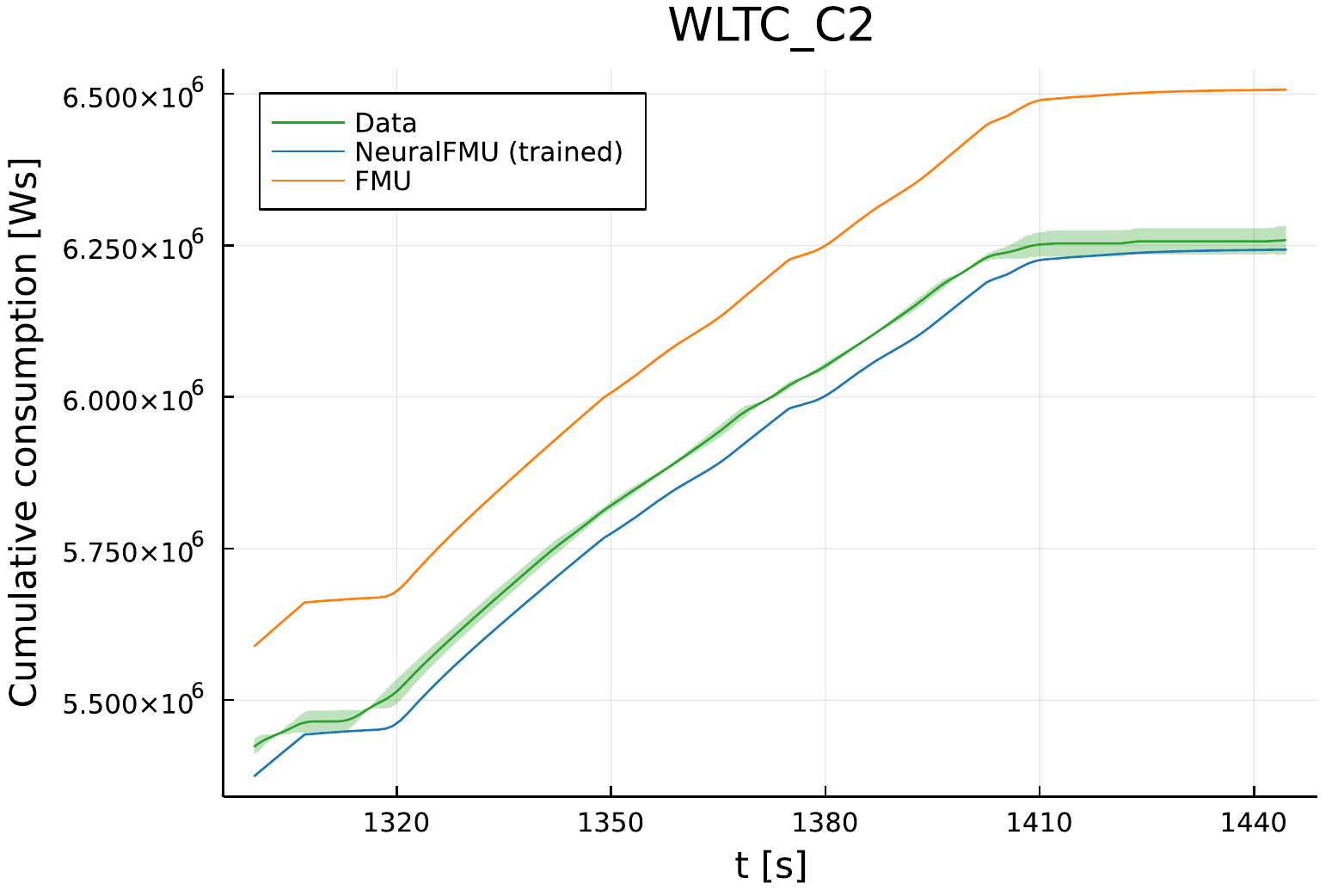}
	\caption{Deviation on the last $10\,\%$ of the simulation trajectory of the NeuralFMU (blue) compared to the original \ac{FPM} as \ac{FMU} (orange) and experimental data mean (green). The unknown \testcycle{} is used for testing. The NeuralFMU prediction is much closer to the data mean than the original \ac{FPM} and predicts a final value inside of the measurement uncertainty.}
	\label{fig:restest10}
\end{figure} 
\begin{figure}[h!]
	\centering
	\includegraphics[width=12cm]{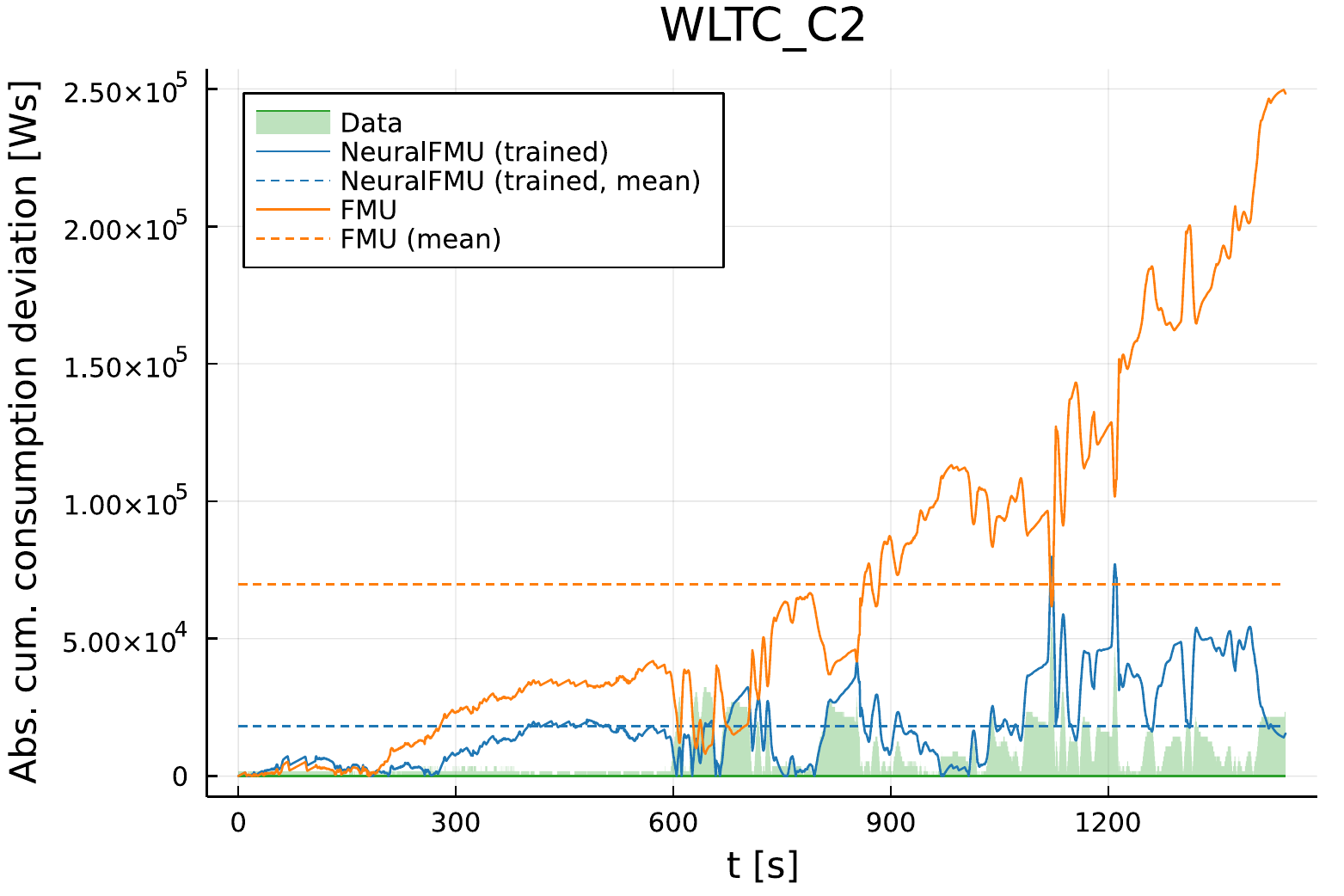}
	\caption{Absolute error of the consumption prediction of the NeuralFMU (blue) compared to the original \ac{FPM} as \ac{FMU} (orange). The unknown \testcycle{} is used for testing. Even if the NeuralFMU solution (blue) does not always lay inside the data uncertainty region (green), it leads to a much better prediction compared to the original \ac{FPM}, that in contrast almost completely misses the data uncertainty area (green, translucent).}
	\label{fig:ertest}
\end{figure}

For training as well as for testing data, the Neural\ac{FMU} solution leads to a smaller \ac{MSE}, maximum error and final error than the solution of the original \ac{FPM}. For the training cycle the Neural\ac{FMU} solution proceeds inside of the measurement uncertainty, which is a great success. Also for testing data, the Neural\ac{FMU} increases prediction accuracy compared to the \ac{FPM}, but leaves the data uncertainty region in some sections. A detailed overview of the training and testing results can be seen in \reftab{trainres}. 
\begin{table}[h!] 
	\caption{Training and testing results with solver \emph{Tsit5} after \numepochs{} training epochs. Errors are calculated against the data mean of two chassis dynamometer runs.\label{tab:trainres}}

		\newcolumntype{C}{>{\centering\arraybackslash}X}
		\newcolumntype{R}{>{\raggedleft\arraybackslash}X}
		\begin{tabularx}{\textwidth}{RRRRRRRR}
			\toprule
			Model & Cycle & MSE [$W^2s^2$] & max. error [$Ws$] & final error [$Ws$]& sim. time [$s$] & solver steps & triggered events\\
			\midrule
			\ac{FMU} & \traincycle{}		& 588.91e8 & 460185.79 & -453614.80 & 10.29	& 110294	& 110247 \\
			Neural\ac{FMU} & \traincycle{}	& 9.30e8 & 63558.61 & -55610.81 & 55.09	& 110301	& 110247		\\
			\ac{FMU} & \testcycle{}			& 89.82e8 & 249693.13 & -248286.71 & 13.16	& 144569	& 144519	\\
			Neural\ac{FMU} & \testcycle{}	& 5.67e8 & 79822.16 & 15463.58 & 69.54	& 144590	& 144519		\\
			\bottomrule
		\end{tabularx}

\end{table}
Because longitudinal dynamics models are often used to predict the cumulative consumption on entire cycles, especially the final value of the solution is important and a key factor of model evaluation. Please note, that this final error of the Neural\ac{FMU} is more than eight times smaller on training data and even 16 times smaller on validation data. Besides the final error, the Neural\ac{FMU} features a much lower \ac{MSE} (factor $\approx 63$ on training, factor $\approx 16$ on testing) and maximum error (factor $\approx 7$ on training, factor $\approx 3$ on testing), but the simulation time increases about five to six times. This is mainly because of the more expensive event-handling inside the hybrid structure and further unused performance optimizations in the prototypical implementation. It can be seen, that the number of events remains unchanged. The number of adaptive solver steps only slightly increases, which indicates that the average system stiffness\footnote{In this article, an \ac{ODE} is considered \emph{stiff} if the adaptive step size of the solver is controlled authoritative by the stability objective, instead of the tolerance objective.} hardly changes. Finally, the training is not converged yet and further training epochs or training on more data (e.g. multiple cycles) may reduce the remaining deviations.

\section{Conclusion}
We highlighted a workflow to allow for hybrid modeling with an industry typical \ac{FPM} in form of a NeuralFMU. Before training such models, a proper initialization is required. Because initialization of NeuralFMUs is not trivial, we suggested three methods with different requirements: \ac{NIPT}, \ac{CCPT} and a topology using \ac{ANN}/\ac{FMU} gates, that makes an initialization routine obsolete. The use of this topology was tested in practice using an industry typical \ac{FPM}, the \ac{VLDM}. This model features multiple challenges like closed-loops and high-frequent discontinuities. The \ac{VLDM} was exported in a format that is common in industrial practice, the \ac{FMI}. On the foundation of the exported \ac{FMU}, a hybrid model was built up and trained on real measurement data from a chassis dynamometer, including typical measurement errors. The model was trained on only a single driving cycle measurement to show that the presented method is capable of making good predictions on very little data. The trained hybrid model was able to make better predictions compared to the \ac{FPM}, on training as well as on testing data. To conclude, the presented workflow and software allows for the re-use of already existing industrial models as cores of hybrid models in the form of NeuralFMUs. NeuralFMUs allow for the data-driven modeling of physical effects, that are difficult to model based on first principle, by using the presented methods and techniques. 

The briefly highlighted open source library \libfmi{} allows the easy and seamless integration of \acp{FMU} into the Julia programming language. \acp{FMU} can be loaded, parameterized, simulated and exported using the abilities of the \ac{FMI} standard. Optional functions like retrieving the partial derivatives or manipulating the \ac{FMU} state are available if supported by the \ac{FMU}. The current library release version is compatible with \ac{FMI} 2.0 (the common version at the time of release) and initial support for \ac{FMI} 3.0 is implemented. The library currently supports \ac{ME}- as well as \ac{CS}-\acp{FMU}, running on Windows and Linux operating systems. Event-handling to simulate discontinuous \ac{ME}-\acp{FMU} is supported. The library extension \libfmiflux{} makes \acp{FMU} differentiable and opens the possibility to setup and train Neural\acp{FMU}. The framework supports proper event-handling during back-propagation whilst training of discontinuous Neural\acp{FMU}. Beside NeuralFMUs, \libfmiflux{} paves the way for other hybrid modeling techniques and new industrial use cases by making \acp{FMU} differentiable in an \ac{AD}-framework. The library repositories are constantly extended by new features and maintained for the upcoming technology progress. Contributors are welcome.

\section*{Funding}
This research was partially funded by the ITEA3-Project UPSIM (\textbf{U}nleash \textbf{P}otentials in \textbf{Sim}ulation) N°19006, see: \url{https://www.upsim-project.eu/} for more information.

\section*{Data availability}
The used software \libfmi{} and \libfmiflux{} is available open-source under \urlfmi{} and \urlfmiflux{}. The data used in the presented experiment (vehicle measurements) as well as the original model are available for download under \url{https://github.com/TUMFTM/Component_Library_for_Full_Vehicle_Simulations}. Further, a tutorial for reconstruction of the presented method focusing on adapting custom use cases will be released soon after this article's publication in the repository of \libfmiflux{}.

\section*{Acknowledgments}
The authors like to thank everyone that contributed to the library repositories, especially our students and student assistants \emph{Josef Kircher}, \emph{Jonas Wilfert} and \emph{Adrian Brune}.

\section*{Abbreviations}
\begin{acronym}[AD]
	\acro{AD}[AD]{Automatic Differentiation}
\end{acronym}

\begin{acronym}[ANN]
	\acro{ANN}[ANN]{Artifical Neural Network}
	\acroplural{ANN}[ANNs]{Artifical Neural Networks}
\end{acronym}

\begin{acronym}[BNSDE]
	\acro{BNSDE}[BNSDE]{Bayesian Neural Stochastic Differential Equation}
	\acroplural{BNSDE}[BNSDEs]{Bayesian Neural Stochastic Differential Equations}
\end{acronym}

\begin{acronym}[CADC]
	\acro{CADC}[CADC]{Common Artemis Driving Cycles}
\end{acronym}

\begin{acronym}[CCPT]
	\acro{CCPT}[CCPT]{Collocation Pre-Training}
\end{acronym}

\begin{acronym}[CS]
	\acro{CS}[CS]{Co-Simulation}
	\acroplural{CS}[CSs]{Co-Simulations}
\end{acronym}

\begin{acronym}[DARN]
	\acro{DARN}[DARN]{Deep Auto-Regressive Network}
	\acroplural{DARN}[DARNs]{Deep Auto-Regressive Networks}
\end{acronym}

\begin{acronym}[FMI]
	\acro{FMI}[FMI]{Functional Mock-up Interface}
\end{acronym}

\begin{acronym}[FMU]
	\acro{FMU}[FMU]{Functional Mock-up Unit}
	\acroplural{FMU}[FMUs]{Functional Mock-up Units}
\end{acronym}

\begin{acronym}[FPM]
	\acro{FPM}[FPM]{First Principle Model}
	\acroplural{FPM}[FPMs]{First Principle Models}
\end{acronym}

\begin{acronym}[HiL]
	\acro{HiL}[HiL]{Hardware in the Loop}
\end{acronym}

\begin{acronym}[ME]
	\acro{ME}[ME]{Model Exchange}
\end{acronym}

\begin{acronym}[ML]
	\acro{ML}[ML]{Machine Learning}
\end{acronym}

\begin{acronym}[MSE]
	\acro{MSE}[MSE]{Mean Squared Error}
\end{acronym}



\begin{acronym}[NEDC]
	\acro{NEDC}[NEDC]{New European Driving Cycle}
\end{acronym}

\begin{acronym}[NIPT]
	\acro{NIPT}[NIPT]{Neutral Initialization Pre-Training}
\end{acronym}

\begin{acronym}[ODE]
	\acro{ODE}[ODE]{Ordinary Differential Equation}
	\acroplural{ODE}[ODEs]{Ordinary Differential Equations}
\end{acronym}

\begin{acronym}[PAC]
	\acro{PAC}[PAC]{Probably Approximately Correct}
\end{acronym}

\begin{acronym}[PINN]
	\acro{PINN}[PINN]{Physics-informed Neural Network}
	\acroplural{PINNs}[PINNs]{Physics-informed Neural Networks}
\end{acronym}

\begin{acronym}[RNN]
	\acro{RNN}[RNN]{Recurrent Neural Network}
	\acroplural{RNN}[RNNs]{Recurrent Neural Networks}
\end{acronym}


\begin{acronym}[SE]
	\acro{SE}[SE]{Scheduled Execution}
\end{acronym}

\begin{acronym}[SSP]
	\acro{SSP}[SSP]{System Structure and Parameterization}
	\acroplural{SSP}[SSPs]{System Structure and Parameterization}
\end{acronym}

\begin{acronym}[VLDM]
	\acro{VLDM}[VLDM]{Vehicle Longitudinal Dynamics Model}
\end{acronym}

\begin{acronym}[WLTC]
	\acro{WLTC}[WLTC]{Worldwide harmonized Light-duty vehicles Test Cycle}
\end{acronym}

\printbibliography

\end{document}